\documentclass[conference]{IEEEtran}
\IEEEoverridecommandlockouts

\usepackage{cite}
\usepackage{amsmath,amssymb,amsfonts}
\usepackage{algorithmic}
\usepackage{graphicx}
\usepackage{booktabs}
\usepackage{textcomp}
\usepackage{xcolor}
\def\BibTeX{{\rm B\kern-.05em{\sc i\kern-.025em b}\kern-.08em
    T\kern-.1667em\lower.7ex\hbox{E}\kern-.125emX}}
    
\begin{document}

\title{Accelerating Reinforcement Learning via Error-Related Human Brain Signals\\

\thanks{This research was supported by the Institute of Information \& Communications Technology Planning \& Evaluation (IITP) grant, funded by the Korea government (MSIT) (No. RS-2019-II190079, Artificial Intelligence Graduate School Program, Korea University).}
}

\author{
    \IEEEauthorblockN{Suzie Kim}
    \IEEEauthorblockA{\textit{Dept. of Artificial Intelligence} \\
    \textit{Korea University}\\
    Seoul, Republic of Korea \\
    sz\_kim@korea.ac.kr}
    \and
    \IEEEauthorblockN{Hye-Bin Shin}
    \IEEEauthorblockA{\textit{Dept. of Brain and Cognitive Engineering} \\
    \textit{Korea University}\\
    Seoul, Republic of Korea \\
    hb\_shin@korea.ac.kr}
    \and
    \IEEEauthorblockN{Hyo-Jeong Jang}
    \IEEEauthorblockA{\textit{Dept. of Brain and Cognitive Engineering} \\
    \textit{Korea University}\\
    Seoul, Republic of Korea \\
    h\_j\_jang@korea.ac.kr}
}

\maketitle

\begin{abstract}

In this work, we investigate how implicit neural feedback can accelerate reinforcement learning in complex robotic manipulation settings. While prior electroencephalogram (EEG)-guided reinforcement learning studies have primarily focused on navigation or low-dimensional locomotion tasks, we aim to understand whether such neural evaluative signals can improve policy learning in high-dimensional manipulation tasks involving obstacles and precise end-effector control. We integrate error-related potentials decoded from offline-trained EEG classifiers into reward shaping and systematically evaluate the impact of human-feedback weighting. Experiments on a 7-DoF manipulator in an obstacle-rich reaching environment show that neural feedback accelerates reinforcement learning and, depending on the human-feedback weighting, can yield task success rates that at times exceed those of sparse-reward baselines. Moreover, when applying the best-performing feedback weighting across all subjects, we observe consistent acceleration of reinforcement learning relative to the sparse-reward setting. Furthermore, leave-one-subject-out evaluations confirm that the proposed framework remains robust despite the intrinsic inter-individual variability in EEG decodability. Our findings demonstrate that EEG-based reinforcement learning can scale beyond locomotion tasks and provide a viable pathway for human-aligned manipulation skill acquisition.

\end{abstract}


\section{INTRODUCTION}
Reinforcement learning from human feedback (RLHF) has emerged as a promising paradigm for aligning agent behavior with human intention~\cite{rlhf, lab7, nipsllm}, enabling autonomous agents to learn complex skills without reliance on manually engineered reward functions. Previous studies have explored the integration of various forms of explicit feedback, such as preference labels or corrective demonstrations~\cite{breadcrumbs, pebble}, and implicit feedback, such as body gestures or facial expressions~\cite{lab17, lab20, empathic}. Recently, non-invasive neural signals like electroencephalogram (EEG)~\cite{lab4, lab5} have emerged as a new form of implicit feedback that can potentially offer a more natural and low-effort communication channel for human–robot collaboration. Among these signals, error-related potentials (ErrPs)~\cite{hri}—cortical responses elicited when humans perceive mistakes—have demonstrated clear utility as evaluative cues within adaptive brain-computer interfaces (BCI) and robotic control systems~\cite{lab6}. Prior studies have shown that ErrP-based feedback can accelerate learning in navigation or locomotion tasks by providing moment-by-moment assessments of behavioral appropriateness~\cite{accelerated, lab1, lab16}, even when explicit reinforcement is unavailable.

Despite these advances, a critical gap remains in understanding whether neural evaluative feedback can scale to more challenging robotic domains. Existing ErrP-driven RL frameworks have largely operated in low-dimensional settings involving discrete action spaces or simple locomotion behaviors, where the control demands are modest and the role of trajectory optimality is limited~\cite{accelerating}. High-DoF manipulation tasks, in contrast, require precise coordination of arm kinematics, continuous control across a multi-joint workspace, and robust handling of obstacles and spatial constraints~\cite{lab18, lab19}. Such tasks amplify both the complexity of credit assignment and the consequences of suboptimal exploration, thereby raising fundamental questions regarding the reliability, stability, and utility of neural feedback under rich sensorimotor dynamics.

To address this challenge, we investigate how implicit neural evaluative signals can accelerate reinforcement learning in a 7-degree-of-freedom robotic manipulation task involving obstacle-rich lifting and placement~\cite{robosuite}. Our approach builds upon the core insight that ErrP-driven evaluative feedback can modulate exploration and guide policy refinement even when the reward landscape is sparse~\cite{Suzie}. The robustness of such neural-feedback systems builds upon foundational progress in machine learning and deep learning methods, including convolutional and recurrent neural networks~\cite{eegnet, lab9}, and various different algorithms that laid the groundwork for modern signal processing and pattern recognition~\cite{lab2, zhang2018cascade, lab3, lab8}. By decoding ErrPs using an offline-trained EEGNet~\cite{eegnet} classifier and mapping these neural responses into scalar shaping rewards, we enable the agent to leverage human internal evaluations during training without explicit human intervention. 

Unlike prior work focused on navigation settings, our study examines how the weighting of neural feedback influences the learning trajectory of a high-dimensional manipulation policy, and whether optimal performance emerges at intermediate feedback strengths.
In this study, we systematically vary the human-feedback weight to characterize how feedback strength modulates learning stability. Beyond episodic return metrics, we introduce manipulation-specific evaluation criteria including path efficiency and collision statistics to capture aspects of motion quality and safety that are critical for robotic manipulators. We further conduct subject-level robustness analyzes in twelve participants, each supplying independent EEG-derived evaluative signals. This enables us to quantify inter-individual variability and determine whether implicit neural feedback remains effective even for subjects with relatively low decoder accuracy.
Across extensive evaluations, we find that small to moderate feedback weights yield consistent improvements in learning speed, trajectory optimality, and collision reduction. Notably, even though sparse-reward agents eventually learn the task, the addition of implicit neural feedback substantially accelerates their convergence by providing supplementary evaluative signals throughout training. 
Overall, this study provides new evidence that neuroadaptive reinforcement learning can scale beyond locomotion, offering a principled pathway toward human-aligned robotic skill acquisition in realistic manipulation settings.


\begin{figure}[!t]
\centerline{\includegraphics[width=\columnwidth]{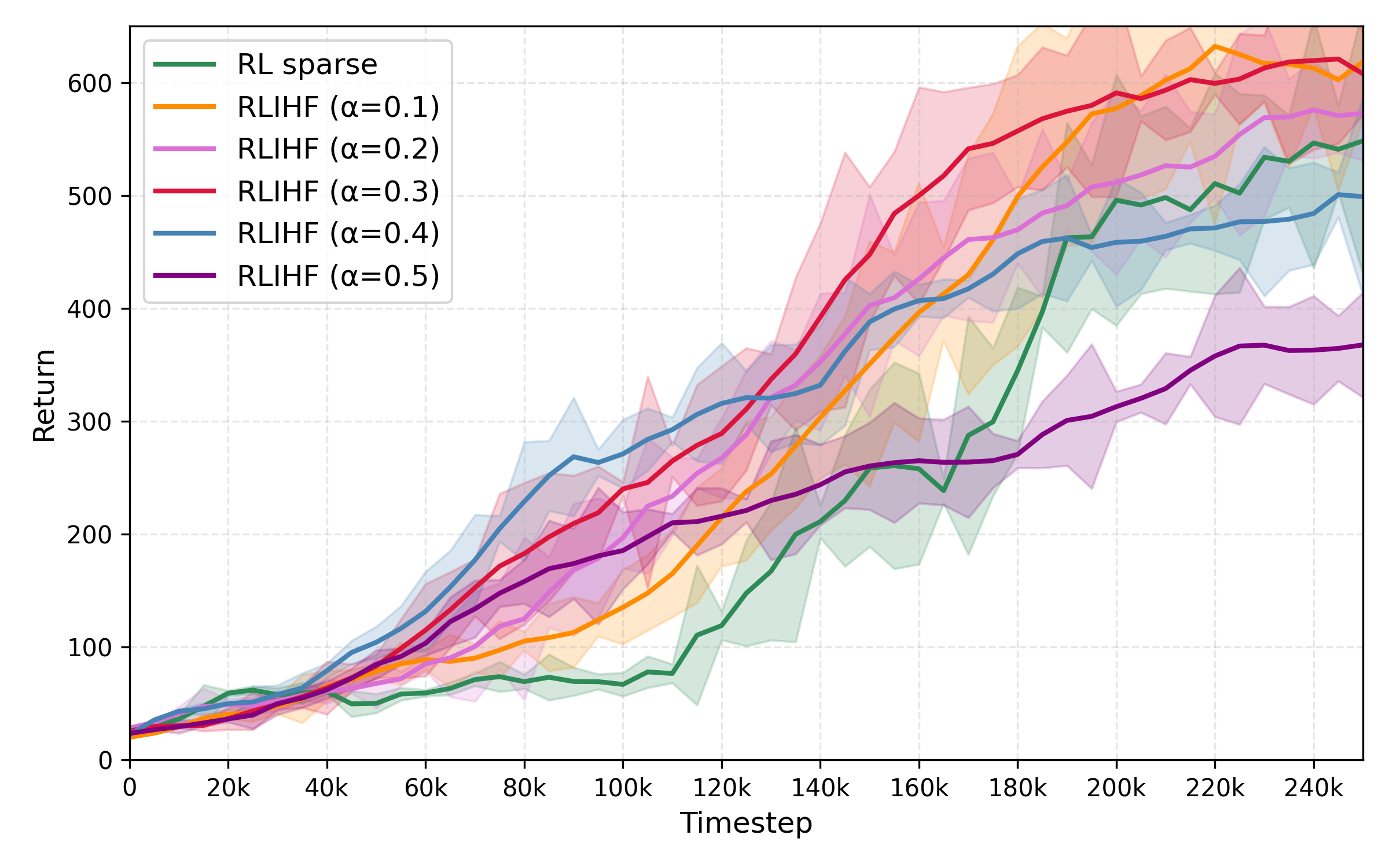}}
\caption{Effect of human-feedback weighting on learning performance. Episodic return curves for sparse RL and RLIHF across different feedback weights ($\alpha$). Moderate weights ($\alpha = 0.1\!-\!0.3$) consistently accelerate learning, while excessively large weights reduce final performance.}
\vspace{-0.5cm}
\label{fig:fig2}
\end{figure}

\section{METHODS}
\subsection{Overview}
Our framework integrates implicit neural evaluative signals into the reinforcement-learning process of a continuous-control manipulation agent. Building upon this foundation, the agent is trained using a reward function that combines sparse task-level reinforcement with an additional human-feedback term derived from decoded ErrPs. To obtain this feedback, raw EEG activity is processed through a pretrained ErrP decoder, which outputs the probability that the human observer would perceive the agent’s current action as erroneous. This probabilistic neural evaluation is then transformed into a scalar reward and integrated with the sparse environment reward, enabling the learning agent to incorporate human internal assessments during policy optimization.

\subsection{ErrP Decoder Pretraining}
To construct a subject-specific neural feedback model, we first pretrain an ErrP decoder using EEG data obtained from twelve participants in a human–robot interaction setting. The raw EEG signals are bandpass-filtered in the 1–20 Hz range to isolate slow cortical dynamics associated with event-related responses, downsampled to reduce redundancy, and re-referenced to minimize channelwise bias. Each recording is segmented into fixed-length epochs time-locked to the moment the participant observes the robot’s action outcome, capturing the characteristic post-stimulus window in which error-related potentials reliably emerge. These uniformly preprocessed epochs form the input samples used for decoder training.

For each participant, an EEGNet-based convolutional model is trained using a leave-one-subject-out (LOSO) strategy. In this scheme, the decoder for a given subject is trained on the epochs obtained from the remaining eleven participants, ensuring subject-independent generalization at test time. EEGNet’s temporal convolutions extract frequency-selective dynamics, while its depthwise spatial filters capture distributed neural patterns associated with error perception. The resulting decoder outputs a continuous probability estimating whether the observed action would elicit an ErrP for that subject. During reinforcement learning, preprocessed epochs are streamed sequentially and passed through the corresponding pretrained decoder, producing temporally aligned neural evaluations that can be directly incorporated into the reward function.

\begin{table}[t]
    \centering
    \caption{Comparison of success rate, path efficiency, and mean collision between sparse and RLIHF reward settings.}
    \label{tab:sparse_vs_hf_fullrange}
    \renewcommand{\arraystretch}{1.2}
    \begin{tabular}{l c c c c c}
        \toprule
        \textbf{Method} & \textbf{Success Rate} & \textbf{Path Eff.} & \textbf{Mean Collision} \\
        \midrule
        RL sparse & 0.22 ± 0.38 & 0.64 ± 0.24 & 2.54 ± 9.49 \\
        RLIHF (\( \alpha=0.1 \)) & 0.33 ± 0.43 & 0.65 ± 0.24 & 2.98 ± 9.91 \\
        RLIHF (\( \alpha=0.2 \)) & 0.30 ± 0.43 & 0.71 ± 0.24 & 3.74 ± 15.79 \\
        RLIHF (\( \alpha=0.3 \)) & 0.37 ± 0.45 & 0.69 ± 0.24 & 2.41 ± 9.03 \\
        RLIHF (\( \alpha=0.4 \)) & 0.26 ± 0.42 & 0.77 ± 0.25 & 3.40 ± 15.72 \\
        RLIHF (\( \alpha=0.5 \)) & 0.14 ± 0.32 & 0.73 ± 0.27 & 1.86 ± 15.17 \\
        \bottomrule
    \end{tabular}
    \label{tab:table}
\end{table}

\begin{figure*}[!ht]
\centerline{\includegraphics[width=\textwidth]{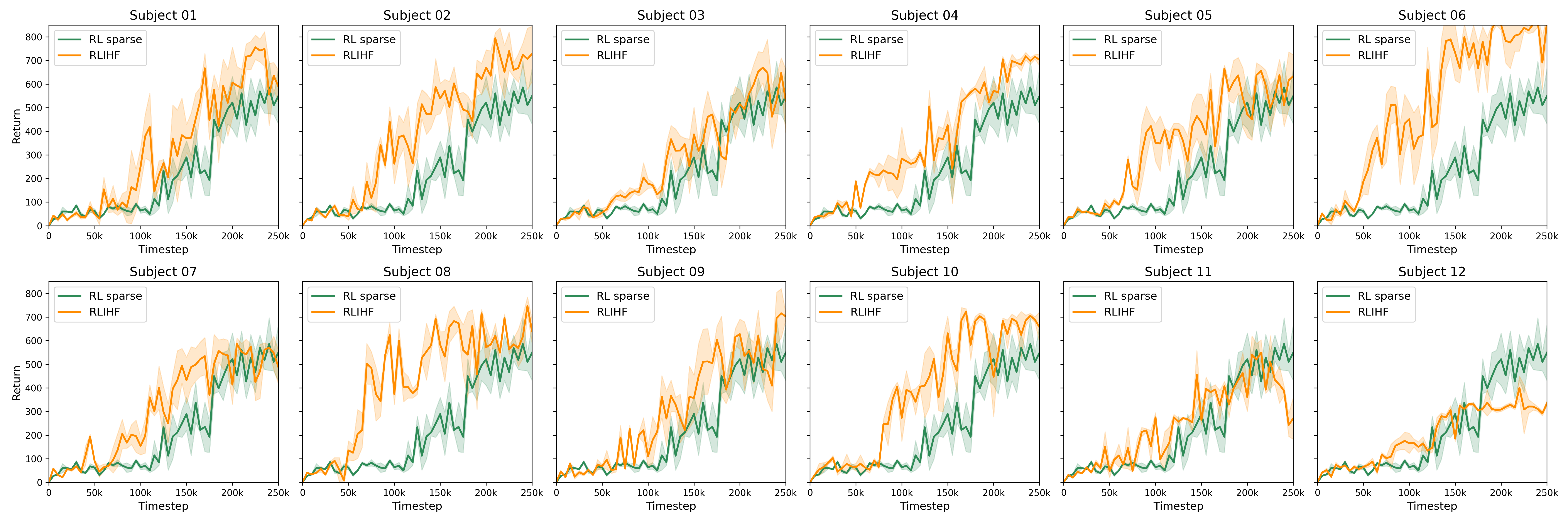}}
\caption{
Cross-subject RLIHF evaluation across 12 participants. For nearly all subjects, RLIHF accelerates learning relative to the sparse-reward baseline, demonstrating robustness to individual variability in EEG decoder accuracy.
}
\vspace{-0.5cm}
\label{fig:fig3}
\end{figure*}

\subsection{Human Feedback Reward Integration}
To incorporate neural evaluative signals into the reinforcement-learning process, we map the decoder’s output probability to a scalar reward term and blend it with the environment’s task-level reward. At each timestep, the pretrained ErrP decoder produces a probability:
\begin{equation}
    p_t \in [0,1],
\end{equation}
indicating the likelihood that the human observer would judge the agent’s most recent action as erroneous. To ensure that uncertain predictions contribute minimally to the reward signal and that highly confident error detections penalize the agent proportionally, we transform this probability using a centered mapping:
\begin{equation}
    r_{\text{hf}}(t) = 0.5 - p_t .
\end{equation}

This formulation assigns positive reinforcement to actions unlikely to evoke perceived error \((p_t < 0.5)\) and negative reinforcement when the decoder detects a high likelihood of error \((p_t > 0.5)\).

The final reward supplied to the RL agent combines this neural feedback component with the sparse task-level reward generated by the environment. Let \(r_{\text{env}}(t)\) denote the sparse environment reward, which reflects only success events and collision penalties. The total reward used for policy learning is computed as a weighted sum:
\begin{equation}
    r_{\text{total}}(t) = r_{\text{env}}(t) + \alpha\, r_{\text{hf}}(t),
\end{equation}
where \(\alpha > 0\) controls the relative influence of neural feedback. This design allows ErrP-derived evaluations to guide policy refinement without overwhelming task-specific reward structure. By integrating neural feedback in this manner, the agent receives dense evaluative signals even when explicit environmental rewards are sparse, thereby accelerating learning while maintaining compatibility with standard off-policy algorithms such as Soft Actor–Critic~\cite{sac}.

\section{EXPERIMENTS}
\subsection{Experimental Setup}
All experiments were carried out in a simulated 7-DoF robotic manipulation environment built on the MuJoCo physics engine and implemented through the robosuite framework~\cite{robosuite}. The robot operated within a cluttered tabletop workspace populated with multiple static obstacles, and was required to reach toward a designated object and transport it to a predefined goal region. To support these evaluations, we employed a modified version of robosuite’s Lift task, in which additional obstacles and a fixed goal area were incorporated to increase task complexity and to more closely reflect real-world manipulation constraints.

To analyze robustness under human-specific feedback conditions, we evaluated our framework using EEG data obtained from a public HRI-ErrP dataset~\cite{hri} collected from twelve human participants observing robot actions. Each subject's data was used to train an independent EEG decoder, and identical reinforcement-learning experiments were repeated for all twelve decoders to examine inter-individual variability. All models were implemented in Python, and policies were trained for 250,000 timesteps using identical training and evaluation protocols across all conditions.

\subsection{Effect of Human-Feedback Weighting on Policy Learning}
To investigate how strongly implicit neural feedback should influence the agent's reward, we varied the feedback weight $\alpha$ and compared learning performance across conditions. As shown in Fig.~\ref{fig:fig2}, low-to-moderate values ($\alpha = 0.1, 0.2, 0.3$) consistently accelerated the growth of episodic return relative to the sparse baseline, demonstrating that neural feedback effectively compensates for insufficient task-level rewards. Among these, $\alpha = 0.3$ achieved the most stable learning curve and the highest final return, suggesting that moderate weighting strikes a balance between providing additional evaluative guidance and maintaining stable policy updates. These effects were further reflected in Table~\ref{tab:table}, where success rate peaked at $\alpha = 0.3$, confirming that well-calibrated human feedback enhances both sample efficiency and task completion.

In contrast, higher weighting values ($\alpha \geq 0.4$) generated a distinct learning pattern: although early gains were observed, performance eventually plateaued or declined during mid-to-late training. This degradation is likely caused by over-amplifying the inherent noise and uncertainty of ErrP-based signals, which can bias the reward toward conservative corrections rather than task-progressing actions. Notably, Table~\ref{tab:table} shows that $\alpha = 0.5$ produced the lowest success rate and reduced return, yet yielded the smallest collision count and lowest collision-occurrence rate. This indicates that an excessively strong influence of neural feedback suppresses risk-taking behavior and shifts the policy toward overly cautious motion planning. While detrimental to overall task performance, this outcome reveals that high-weighted feedback naturally promotes safety-oriented behaviors, suggesting potential applications in risk-sensitive or human-aligned robotic control.

\subsection{Cross-Subject Robustness of Implicit Neural Feedback}
To assess whether the performance gains observed with implicit human feedback generalize across individuals with heterogeneous EEG characteristics~\cite{lab10}, we conducted a subject-level robustness analysis using the best-performing feedback weight, $\alpha = 0.3$. For each of the twelve participants in the EEG dataset, the corresponding EEGNet decoder was used to generate neural evaluative signals during policy training, and each experiment was repeated across five random seeds. The resulting episodic return curves, shown in Fig.~\ref{fig:fig3}, reveal a consistent pattern across nearly all subjects: RLIHF accelerates learning relative to the sparse-reward baseline, yielding faster early-stage convergence and higher returns in the later stages of training. Notably, this trend persists even for participants whose EEG classifiers exhibit only moderate decoding accuracy, indicating that ErrP-derived feedback---despite its noise and subject dependence---still provides a sufficiently informative signal to guide policy improvement. Taken together, these findings confirm that implicit human feedback reliably accelerates reinforcement learning, even under substantial inter-individual variability in neural signal quality.

\section{CONCLUSIONS}
This work demonstrates that implicit neural evaluative signals, specifically error-related potentials decoded from EEG, can reliably accelerate reinforcement learning in high-DoF robotic manipulation tasks. By integrating a centered, probability-based human-feedback reward with sparse environmental reinforcement, the proposed framework provides dense evaluative guidance that substantially speeds up policy convergence while preserving stability in continuous-control settings. Our $\alpha$-sweep analysis revealed that small-to-moderate feedback weights consistently produced faster return growth, improved trajectory optimality, and reduced collisions, confirming that neural feedback acts as an effective catalyst for learning in otherwise sparse and challenging reward landscapes. Moreover, cross-subject evaluations using twelve independently trained decoders showed that this acceleration effect persists despite substantial inter-individual variability in EEG signal quality. Even subjects with moderate decoder accuracy contributed sufficiently informative evaluative cues to guide policy improvement. These findings collectively indicate that implicit neural feedback not only scales to high-dimensional manipulation but also offers a practical pathway toward efficient, human-aligned skill acquisition.



\bibliographystyle{IEEEtran}
\bibliography{reference}

\end{document}